\documentclass{article}

\usepackage{arxiv}

\usepackage[utf8]{inputenc} 
\usepackage[T1]{fontenc}    
\usepackage{hyperref}       
\usepackage{url}            
\usepackage{booktabs}       
\usepackage{amsfonts}       
\usepackage{nicefrac}       
\usepackage{microtype}      
\usepackage{graphicx}
\usepackage{xcolor}
\usepackage{subcaption}
\usepackage{multirow}

\title{Ensemble of Convolution Neural Networks on Heterogeneous Signals for Sleep Stage Scoring}

\author{
  Enrique Fernandez-Blanco \thanks{Corresponding Author} \thanks{
  Alternative Affiliations: \newline
  CITIC. University of A Coruna. A Coruña, 15071, Spain \newline
  INIBIC, Complexo Hospitalario Universitario de A Coruña, A Coruña, 15006, Spain} \\
  Faculty of Computer Science\\
  University of A Coruna\\
  A Coruna, 15071, Spain\\
  \texttt{enrique.fernandez@udc.es} \\
   \And
  Carlos Fernandez-Lozano \footnotemark[2]\\
  Faculty of Computer Science\\
  University of A Coruña \\
  A Coruña, Spain \\
  \texttt{carlos.fernandez@udc.es}\\
  \AND
  Alejandro Pazos \footnotemark[2]\\
  Faculty of Computer Science\\
  University of A Coruna\\
  A Coruna, Spain\\
  \texttt{alejandro.pazos@udc.es} \\
  \And
  Daniel Rivero \footnotemark[2]\\
  Faculty of Computer Science\\
  University of A Coruna\\
  A Coruna, Spain\\
  \texttt{daniel.rivero@udc.es} \\
}

\begin{document}
\maketitle

\begin{abstract}
Over the years, several approaches have tried to tackle the problem of performing an automatic scoring of the sleeping stages. Although any polysomnography usually collects over a dozen of different signals, this particular problem has been mainly tackled by using only the Electroencephalograms presented in those records. On the other hand, the other recorded signals have been mainly ignored by most works. This paper explores and compares the convenience of using additional signals apart from the electroencephalograms. More specifically, this work uses the SHHS-1 dataset with 5,804 patient which contains an electromyogram recorded at the same time as two electroencephalograms. To compare the results, first, it has been evaluated the same architecture with different input signals and all their possible combinations. These tests show how, using more than one signal especially if they are from different sources, improves the results of the classification. Additionally,  the best models obtained for each combination of one or more signals have been used in ensemble models and, its performance has been compared showing the convenience of using these multi-signal models to improve the classification. The best overall model, an ensemble of Depth-wise Separational Convolutional Neural Networks, has achieved an accuracy of 86.06\% with a Cohen's Kappa of 0.80 and a $F_{1}$ of 0.77. Up to date, those are the best results on the complete dataset and it shows a significant improvement in the precision and recall for the most uncommon class in the dataset. 
\end{abstract}


\small{
\section*{Highlights}
\begin{itemize}
    \item Processing heterogeneous signals simultaneously in a Separable CNN and explore the advantages of this approach
    \item Importance of combining various patients in the train batches
    \item Explore the ensemble of multi-signal models and its differences with ensemble of single-signal models
    \item Establish difference between the EEGs signals present on the dataset.
\end{itemize}
}

\section{Introduction}

According to the World Health Organisation (WHO), sleep disorders can be behind other major diseases \cite{world2015global}, such as depression, stress, heart problems, diabetes or early Alzheimer \cite{stranges2012sleep}. With the current pandemic spreading all over the world, the problems to obtain a good quality sleep have been significantly increased \cite{jahrami2021sleep}, resulting in 40\% of the population who claim problems to sleep a straight night. As a  consequence, over time, the activity in the sleeping units in most of the hospitals has registered a significantly increase \cite{ford2014trends}. That increment in the activity goes along with an increment in the number of polysomnographies (PSG) to be processed by the physicians. PSGs are records of simultaneous signals that can be registered during sleep, such as Electrocardiograms, Electroencephalograms (EEGs), Electromyograms (EMG), breath or movement records, etc. These records taken during periods between 6 to 24 hours are the primary tool of the specialist to study, identify and treat the sleep disorders.

To process those records, the physicians have to label the signals identifying the different stages based mainly on their expertise. That labelling process usually follows one of the two guidelines established by the community known as Rechtschaffen \& Kales (R\&K) \cite{rechtschaffen1968manual} and the American Academy of Sleep Medicine (AASM) \cite{berry2012rules}. The main problem with those guidelines is the absence of a clear set of features to identify the different sleep stages. Therefore, the result is a difficult situation prone to make mistakes due to fatigue or monotony\cite{boashash2016automatic}, where each specialist has to process a significant quantity of data in a highly time-consuming process.

In order to solve this situation, different approaches have been made to automate this process. However, this process is particularly tough for any machine learning approach due to the unreliability of the ground truth. For example, in \cite{whitney1998reliability} the physicians involved in the study only agree in the 81\% of labels for the same PSGs, while in \cite{Norman2000Interobserver} it was set around 84\%. Therefore, there is much room for personal interpretation bound to this process, which drives to inconsistent ground truth.

Several techniques have been used to deal with this issue especially focusing on using different automatic feature extraction techniques, such as Genetic Programming \cite{fernandez2015hybrid}. However, Convolutional Neural Networks (CNN) has become a hot topic when, due to their capability to extract information, the specific features to solve a problem are unknown. In fact, this kind of technique has been used previously to tackle the same problem \cite{sors2018convolutional, fernandez2020convolutional, Pathak2021STQS:Scoring}. However, independently of the approach, all the previous works share in common that the techniques only deals with EEGs, while the remaining of the PSG is not processed or ignored in order to simplify the processing. Only in \cite{Pathak2021STQS:Scoring}, other signals are used but they are treated separately before combining the information in the last steps. The tendency of the later approach is to generate silos of information instead of combining the information extracted from the different sources.

Taking as frame the use of Deep Learning (DL) Models to process the information contained in the signals of a PSG, this paper explores which is the impact of the inclusion of the EMG signal in the process of scoring one of the biggest public datasets of PSGs known as SHHS-1 \cite{quan1997sleep}. The idea is to measure if there is any improvement in the convolutional network models and if the CNNs are able to extract information that combines heterogeneous signals in order to improve the performance, as other works have made with only the EEGs \cite{fernandez2013classification, fernandez2020convolutional}. Additionally, with the best models obtained for each combination of signals, different ensembles have been constructed and compare to the ensemble of the single-signal-input models in order to identify if there is additional information extracted by the multi-signal-input models. The result is a model with comparable results to the current state-of-the-art but higher precision and recall for the less common class which also requires a fraction of the parameters of \cite{sors2018convolutional}.This last point also benefits not only a significative reduction in complexity but in training time.

The outline of the paper is organised as follows: section \ref{sec:stateoftheart} present a review of related works with special attention to the most modern ones which have used part of the same dataset. Section \ref{sec:materialsand methods} presents not only the architecture used as base, but a summary of the key elements of convolutional neural networks and a description of the dataset and the signals used on the tests. Those tests and the analysis of the results could be found in \ref{sec:test}. Finally, section \ref{sec:conclusions} and section \ref{sec:future} present the conclusions and the future lines to work, respectively.

\section{State of the art}\label{sec:stateoftheart}

Over the years, the problem of scoring the signals from PSG has received uncountable efforts to improve the extraction of features. Particularly, the one which has especially hogged the light among any other has been the processing of EEGs, while other uses have been less prominent.

Focusing on EEGs processing, many researchers have attempted different approaches to improve the automation of the sleep stage scoring. Although, the number of work is significantly lower if the spotlight is over those that performed an automatic extraction of features. Probably, \cite{berthomier2007automatic} is the first worth to be mention in this sense, where the authors combine Fuzzy Logic with an iterative method to identify the sleeping stages by extracting the features from the signal timeline. Another work worth to be mention that also kept the integrity of the signals' timeline is \cite{liang2012rule}, which applies a decision tree on the raw signal. This point is important due to it could help physicians to understand the decisions of the system. On the other hand, others works have preferred to apply different transformations to the signal, such as the calculation of the entropy together with a Linear Discriminant Analysis (LDA) \cite{liang2012automatic} or the use of wavelets\cite{hassan2017automated}. Going along with the line, \cite{hsu2013automatic} proposes to calculate a combination of energies of the signal and used an Artificial Neural Network (ANN) to perform the classification. Also, closely related, there are those works that do not use the signal but they focus on high-level features such as statistical features \cite{hassan2015classification}, power spectral density \cite{ronzhina2012sleep}, graph theory features \cite{zhu2014analysis} or moment features \cite{hassan2015automatic}.

However, the ones that have focused the attention on recent times have been those related to any Deep Learning approach, such as, Random Belief Networks \cite{hinton2009deep} or Convolutional Neural Networks \cite{lecun1998gradient} due to their capabilities to identify the features to perform the classification. For example, \cite{tsinalis2016automatic} uses Autoencoders to score a well-known dataset of 20-patient \cite{kemp2000analysis}. Supratak et. al. \cite{supratak2017deepsleepnet} proposed an alternative architecture named DeepSleep which was used with the same dataset but also tested with another very common one \cite{o2014montreal}, although at the cost of a very big architecture with a dual pipeline. Trying to address this issue, Fernandez-Blanco et. al. proposes the use of Depth-wise separable convolutional neural networks (DS-CNN). This latter work goes along the line of shown in \cite{fernandez2020convolutional} which points out to an advantage of processing several signals at the same time with a Convolutional Neural network. The results were equivalent to architectures 1000 times bigger in the number of parameters \cite{Biswal2018Expert-levelNetworks, sors2018convolutional} which also use the same dataset, known as SHHS1 \cite{quan1997sleep}. This dataset is up to date the biggest collection of PSG recorded by National Sleep Research Resource with records of 5804 patients.  

The main problem of any of the previously mention works is that, without exception, all of them use only one or two at most of the data signals present in the PSG, and they are always EEGs. However, in any PSG, there are more signals than the EEGs and its addition can show benefits in terms of performance. Therefore, recently, some works have tried to explore the incorporation of alternative signal into the classification. Most of those approaches, e.g. \cite{chambon2018deep,Paisarnsrisomsuk2018DeepTime-Signals,Yildirim2019ASignals}, are very limited attending to its extension with a very small number of patients. The first work which is worth to be mention is \cite{Pathak2021STQS:Scoring}, which focused on using CNN to process each one of the types of signal in the study and combine the result with a  bidirectional long short-term memory in order to focus on the temporal component of the signals, but the separation of the signals in 3 different CNNs make to the network impossible to extract features combining the information of both sources. On the other hand, it is also worth to be mention \cite{Alvarez-Estevez2020AddressingScoring}. This work explores the same database among others with an ensemble approach by developing different CNN models trained with several sources. However, it also presents the same problem as \cite{Pathak2021STQS:Scoring} because the models used on the ensemble are always developed on a single type of signal and then combined in an ensemble. Additionally, this work does not use the complete dataset, but it takes a subset of patients which makes their results very difficult to compare. This work is also going to incorporate a different kind of signal in the classification, Electromyogram (EMG). However, as the main difference with the previously mentioned ones, in this work, the models are going to explore the active combination of the signals in a single model. This allows the network to extract features that can combine both signals in order to obtain more information. Additionally, and using the whole SHHS-1 dataset, it is going to measure the impact of using those new heterogeneous-signal models in different ensembles in order to complement the information extracted by another model. 

\section{Materials and methods}\label{sec:materialsand methods}

\subsection{Convolutional Neural Networks}
Even though Convolution in Neural Networks (CNNs) were firstly proposed by Fukushima in 1982 \cite{fukushima1982neocognitron} and revisited by Yann LeCunn \cite{lecun1998gradient} in 1998, it was not until 2012 that they could get out of the laboratory and used in real-world problems, mainly due to the modifications in the calculation of the gradients proposed by Hinton \emph{et. al.} in \cite{krizhevsky2012imagenet}. Since that key point in their history, CNNs have become an important step forward in many knowledge areas by becoming the state-of-the-art to solve many problems. The success of this approach is based on using a hierarchy of layers which neurons take as input a piece of spatial-close information. Accordingly, each neuron on a convolutional layer receives a different piece of information similar to use a sliding-windows on a signal, while the weights are the same for every single neuron on a layer. So, the result is a function which multiplies an input feature map $X^{(l-1)}$ by a set of learnable filters $W^{(l)}$ while adds a set of biases $b^{(l)}$. The final result  comes from the application of a transfer function $g$ like in Eq.\ref{eq:convolution}.

\begin{equation}
    X^{(l)}=g^l(X^{(l-1)}*W^{(l)} + b^{(l)})
\label{eq:convolution}
\end{equation}

When the process described by Eq. \ref{eq:convolution} is repeated in several layers stacked one over another, the results are the extraction of more general information in each layer of the networks based on the features identified by the previous one \cite{lecun2015deep}. Therefore, CNNs perform the extraction of the features of a signal or image according to the deep or number of convolutional layer they are composed of. Those extracted features are going to be used as the input of the last part of the network which is the one determining if the problem to solve is a regression or a classification. As possible examples of this last part of the network, it can be a Softmax function applied on the outputs, a support vector machine, a multilayer perceptron, or any other machine learning approach.

Although this kind of solution has been mainly applied to image processing\cite{taigman2014deepface,russakovsky2015imagenet}, there are also examples of successful applications in signal processing, for example, in natural language processing \cite{hinton2012deep} or human voice recognition \cite{chan2016listen}. However, independently of the problem, convolution has always been a very expensive operation in time in resources. In order to tackle this problem, in 2017, the Depth-wise Separable Convolutional Neural Networks (DS-CNN) \cite{chollet2017xception} were proposed as a modification, which can dramatically reduce the requirements by severely cutting down the number of parameters.  

\begin{figure}
    \centering
    \includegraphics[width=9cm]{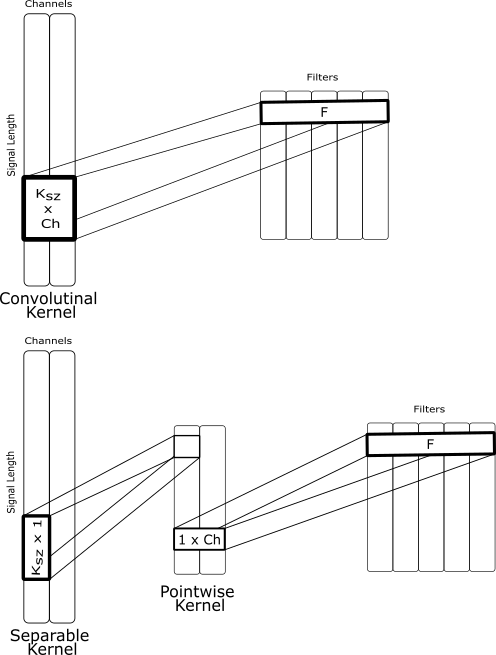}
    \caption{Comparison between a convolution on a couple of signals (top) and its equivalent Depth-wise Separable Convolution (bottom)}
    \label{fig:separableConvolution}
\end{figure}

To do that, this type of convolution combines two simple convolutions. As it is shown in Fig. \ref{fig:separableConvolution}, first, it applies a spatial convolution which is performed on each input channel to extract a small number of features. For example, in the figure,  the separable kernel applied is $K_{sz}\times 1$ to extract 1 feature at a time for each signal. The second step is a convolution performed as a point-wise kernel, i.e. $1 \times Ch$ convolution, which combines the information of the different channels and projects that information in a new space to extract F features.The result is exactly equivalent to the application of a convolutional kernel. However, By closely looking to equations Eq.\ref{eq:OperationsInConvolution} and Eq.\ref{eq:OperationsInSepConvolution}, the benefits are evident in time and memory costs when the number of parameters for each type of convolution is compared.

\begin{equation}
    \#Convolutions = Ch * K_{sz} * (S - K_{sz}) * F
\label{eq:OperationsInConvolution}
\end{equation}

\begin{equation}
    \#SeparableConv = Ch* K_{sz} * (S - K_{sz})  + Ch *(S-K_{sz}) * F
\label{eq:OperationsInSepConvolution}
\end{equation}

Eq.\ref{eq:OperationsInConvolution} represents the number of operations in a traditional convolutional operation, with $Ch$ being the number of channels, $K_{sz}$ is the size of the kernel to be applied, $F$ the number of filters and $S$ is the length of the signal. If this operation is compared to the number represented in Eq. \ref{eq:OperationsInSepConvolution}, representing the number of operations in an equivalent separable convolution, the results is the ratio shown on Eq.\ref{eq:RatioConvolution}. This relationship points out to a reduction which is inversely proportional to the size of the kernel and the number of Filters. For example, suppose that our network uses a kernel of size $11\times2$, as it does, and it extracts 20 features or applies 20 filters with each convolution. The result of the relation in Eq. \ref{eq:RatioConvolution} for that kernel will be $\frac{1}{11\times2} +\frac{1}{20} \approx 0.095$, that is a massive 90.5\% reduction in the number of needed gradients need for a single layer and in the number of weights required. Consequently, the result is a cheaper operation while it has an equivalent result. 

\begin{equation}
   \frac{\#SeparableConv}{\#Convolutions} = \frac{1}{K_{sz}} + \frac{1}{F}
\label{eq:RatioConvolution}
\end{equation}

\subsection{Evaluation Criteria}
In this work, the dataset was randomly split into 3 subsets, training, validation and test, following 0.7, 0.1 and 0.2 ratios, respectively. The reason for such division was to control the possible overfitting of the model and allow a fair comparison with other works. In fact, the model to be chosen to perform the test would be the one with the lowest validation value through the training process. With this point in mind, three have been the main measures analysed to determine the performance of the model: accuracy, Cohen's Kappa and $F_{1}$-$score$. All of them redeem from the confusion matrix, the accuracy represented in Eq.\ref{eq:Accuracy} measures the raw performance of the model by finding the ratio between the number of true positives (TP) and true negative (TN) instances over the total number of instances, i.e, the number of true positives, true negatives, false positives (FP) and false negatives (FN).
\begin{equation}
    Accuracy = \frac{TP+TN}{TP+FP+FN+TN} 
    \label{eq:Accuracy}
\end{equation}

On its own, Cohen's Kappa estimates the agreement between the algorithm and the technicians by removing the bias of any of the classifiers. By following the Eq.\ref{Eq:Kappa}, this estimation excludes the chances of random agreement. In that formula, $p_0$ represents the observed agreement while $p_e$ is the probabilities of a chance agreement. Subtracting this latter probability, the result is a ratio of the similarity in the decisions between two classifiers.
\begin{equation}
    \kappa = \frac{p_0 - p_e}{1 - p_e}
    \label{Eq:Kappa}
\end{equation} 

Finally, represented in Eq.\ref{eq:F1}, $F_{1}$\textit{-score} combines precision (or positive predictive value, PPV) and recall (or true positive rate, TPR) in a single measure through a geometric average. PPV represents the ratio between the true positives (TP) among all the cases labelled as positive by the model, while TPR represents the positive cases identified among the total number of positives in the ground truth. It is worth mentioning that when dealing with multiclass, $F_{1}$ can be calculated attending to different approaches which can significantly modify its result. Those approaches usually received the name of micro, macro and weighted. In this work, it has always been used the macro approach where The $F_1$ is calculated for each class and the average of these ones is presented. 
\begin{equation}
    F_1\textrm{\textit{-score}} = 2*\frac{PPV*TPR}{PPV+TPR} 
    \label{eq:F1}
\end{equation}

\subsection{Dataset}\label{sec:Dataset}
Data used in this work comes from a multi-center cohort study known as Sleep Heart Health Study (SHHS) \cite{quan1997sleep}, which was carried out from 1995 until 2010 by American National Heart Lung and Blood Institute. The initial objective was to determine the relationship between sleep-disorders and high-risk cardiovascular issues. The study consists of a series of polysomnographies each of which contains two EEG channels (C4-A1 and C3-A2), two electrooculograms (EOG) channels, one el; electromyogram (EMG) channel, one electrocardiogram (ECG) channel, two inductance plethysmography channels (thoracic and abdominal), a position sensor, a light sensor, a pulse oxymeter, and an airflow sensor. The study contains the records from 5,804 patients recorded from 6 to 10 hours. From all those signals, this paper is going to focus its attention on three of those signals, the two channels of EEG and the EMG signal. The resulting subset of dataset used contains 79 Gbytes of information.

\begingroup
\setlength{\tabcolsep}{10pt} 
\renewcommand{\arraystretch}{1.5} 
\begin{table}[ht]
    \caption{Conversion between R\&K and AASM guidelines }\label{tab:sections}
    \centering
    \scriptsize
    \begin{tabular}{l c ccccccc}
        \textbf{Guideline} & \multicolumn{8}{c}{\textbf{Sleeping Stages}}\\[2ex]
        R\&K & & Awake & S1 & S2 & S3 & S4 & REM & Unknown\\[1ex]
        AASM & & Awake & N1 & N2 & \multicolumn{2}{c}{N3} & REM & - \\
    \end{tabular}
\end{table}
\endgroup

It should be highlighted that this dataset is divided into two different subsets: SHHS-1 and SHHS-2, corresponding to two different visits of the same patient cohort. Unfortunately, in this work due to the use of the EEGs, only the SHHS-1 was usable. While SHHS-1 has both EEG channels recorded at 125Hz, SHHS-2 has records sampled between 125Hz and 128Hz with no clear pattern, this is a problem to use any machine learning technique which is usually not prepared to a change on the sample ratio. Therefore, due to its homogeneity, SHHS-1 was the one used as in any other work of the literature. Each PSG in this dataset has been manually scored by a single technician according to R\&K scoring rules \cite{rechtschaffen1968manual}. Therefore, the dataset used contains records from 5804 patients which were labelled on 30s epochs in several sleep stages: Awake, S1, S2, S3, S4, REM and Unknown. These labels were later adapted to the new standard called AASM standard \cite{berry2012rules} because the SHHS was collected before this standard was set. This adaptation was made by following the guidelines also mentioned in \cite{berry2012rules}. The result was a label between Awake, N1, N2, N3, and REM for each epoch as it is shown in Table \ref{tab:sections}.

\subsubsection{Electroencephalograms}
In order to study the brain, electroencephalography is one of the very few non-invasive techniques that can be used. Recorded in electroencephalograms, these tests register the alternations that the brain produces due to its electrical activity over time. It has been widely used in the study and treatment of different clinical problems and diseases, such as diagnosis of epilepsy \cite{fernandez2013classification}, depth of anaesthesia \cite{esmaeilpour2016analyzing}, or sleep disorders \cite{sors2018convolutional}.

\begin{figure}[h]
\begin{subfigure}{0.5\textwidth}
\centering
\includegraphics[scale=0.55]{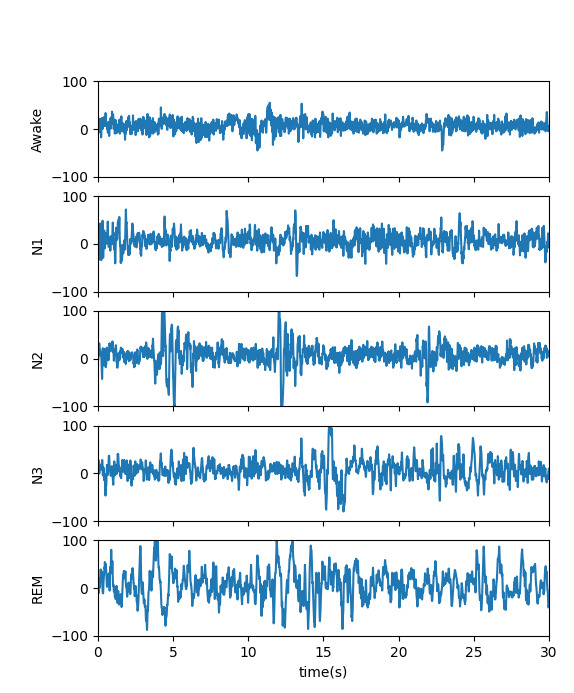} 
\caption{EEG}
\label{fig:exampleEEG}
\end{subfigure}
\begin{subfigure}{0.5\textwidth}
\centering
\includegraphics[scale=0.55]{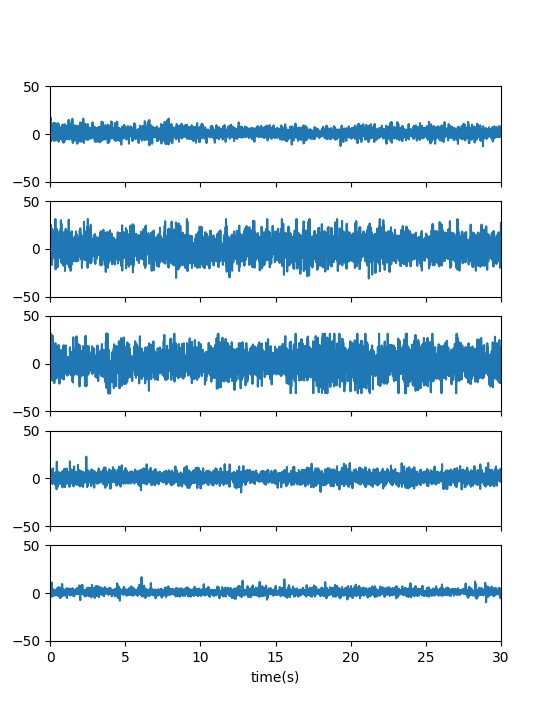}
\caption{EMG}
\label{fig:exampleEMG}
\end{subfigure}

\caption{Samples for both types on signal of each labeled stage in SHHS-1 dataset}
\label{fig:exampleSHHS-1}
\end{figure}

To capture these signals, a set of electrodes placed on the patient's scalp is used. These electrodes are placed by following the standard 10/20 \cite{oostenveld2001five} and calculating the difference between the potential of two electrodes. By naming the positions of the pair of electrodes, a channel can be easily identified, for example, in this work channels C4-A1 and C3-A2 are going to be used because those 4 points were the ones used when the dataset was recorded. 

As one of the major challenges of working with this type of signal, the level of noise can be named as one of the main issues. The origin of this noise come from the preamplification required in the measured potential, this potential in the scalp has an amplitude between 10 µV and 100 µV, which has to be amplified between 1,000 and 10,000 times in order that the instruments can register it. This required amplification results in particular complex signals with many artefacts and false positives which is also non-stationary, as shown in Fig. \ref{fig:exampleEEG}.

\subsubsection{Electromyograms}
Another source of information for the technicians is the records of the muscular activity. Electromyograms (EMG) are the graphical representation of that muscular activity which is recorded by registering the electrical impulse which can be read in on the skin and which are responsible for the movement of the muscle. The record is usually performed by a gold patch which is positioned over the skin and the muscle which activity is required to be recorded. As most of the biological signals, due to the low potential of the signal to register, the records has to preamplified the signal in order to be able to record the activity, however, this results in very noisy signals. For example, the sensor used to record the EMG in the SHHS-1 had a sensibility of 62.5 µV, therefore it needs to be preamplified as well as the EEGs while the device sampled the signal at 125Hz.

Most of the PSG contain at least one of these signals due to its low cost and invasion in the patient. For example, in the particular case of the SHHS-1, there is a single signal which was recorded putting a sensor on the lower chin of the patient in order to register any movement of the jaw. An example of these records can be seen in Fig. \ref{fig:exampleEMG}. 

\subsection{Architecture used on the tests}\label{sec:Architecture}
\begin{figure}[h]
    \centering
    \includegraphics[scale=0.65]{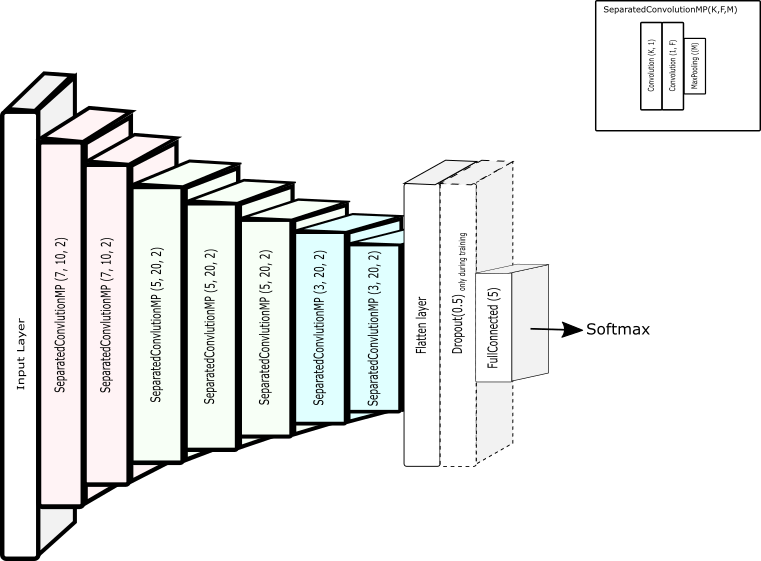}
    \caption{Architecture of the individual models to solve the problem and perform the tests}
    \label{fig:Architecture}
\end{figure}

Tests in this paper have used the architecture proposed in \cite{Fernandez-Blanco2020EEGStage}. This architecture is based on depth-wise separable convolutional layers to solve the problem of scoring the different sleep stages. As it can be seen in Fig. \ref{fig:Architecture}, the proposed schema uses as feature extractor a series of convolutional neural networks, while the classifier is reduced to a single fully-connected layer.

This schema received 1, 2 or 3 signals as inputs which can be either EEG or EMG from the SHHS-1. Any of those signals, which were recorded simultaneously at 125Hz each, were labelled in intervals of 30 seconds as it is established in the R\&K and AASM guidelines. Consequently, a label is available for every 3750 samples in the input signal, which are named sections. As it was established in \cite{Fernandez-Blanco2020EEGStage}, the networks receive 2 additional sections before and after the section that is wanted to label of 1 or more of the previously described signals. 

Independently of the number of signals in the input, the schema of the features extractor has been set as a succession of depthwise separable convolutional layers followed by \emph{MaxPooling}. The particular combination of those 2 convolutional layers and the max-pooling layer is represented in Fig. \ref{fig:Architecture} by blocks named \emph{SeparatedConvolutionMP}. Those blocks define the size of the kernel($K$), the number of features($F$) and the max-pooling size ($M$) as arguments in that precise order. In the proposed schema, Kernels sizes run from 7 to 3. Those sizes go along with the achievements of \cite{sakhavi2018learning}, which points to 7 as the best size to extract information from the brain. The authors of \cite{Fernandez-Blanco2020EEGStage} point to the reduction in the size as the result of an empirical process that tries to increase the pressure in the parameters to improve the generalisation.

It may be highlighted that this particular combination of layers increases the pressure on the extracted features. That statement is supported, first, because each couple of layer which perform a separable convolution uses fewer parameters so that parameters are more important to represent the solution space. Second, a reduced number of filters was used, in this architecture, each layer extract between 10 to 20 filters less than other works which count the filters starting from 100. Finally, the \emph{MaxPooling} layers reduce the number of features by half in each layer. Let's compare the inputs and outputs of the first and last layer of the extractor in order to make an idea of the pressure. Assuming that 3 signals and 5 sections for signal are used as input, there are 56,250 inputs while in the input of the classifier with the proposed architecture there are only 2,860 input values. Therefore, in the last layer, each feature would represent approximately 19.67 features of the first-layer input. 

After this, those features are flattened before going to a full-connected layer of 5 elements which performs a Softmax function. It is worth to be mention that, before this last layer and only during training, there is a Dropout function applied to the connections between the output of the convolutional phase an the full-connected layer. With a probability of 0.5 to drop each of the connections, it is used to improve the generalisation and prevent overtraining.

\subsection{Ensemble methods}\label{sec:ensemble}
\begin{figure}[h]
    \centering
    \includegraphics[height=5cm]{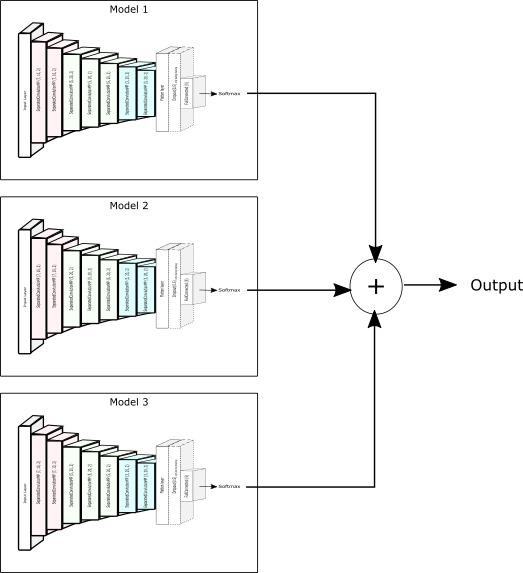}
    \caption{Schema of how the stacking ensemble was performed}
    \label{fig:EnsembleArchitecture}
\end{figure}

In recent times, one of the most successful approaches to classification and regression problems has been a set of techniques called ensemble models. These models rely on the strength of a set of weak or not as capable models which have been trained for the same task. Therefore, the idea behind the ensemble of the models is the combination of the outputs of those weaker models into a more robust one. To develop that combination there are many approaches, although the most common are: bagging\cite{Efron2003SecondBootstrap}, boosting\cite{Freund1997ABoosting} and stacking. The last one is simply to establish a policy in order to get an answer from the particular answers of each model, such as max voting or averaging the output. In this work, as can be seen on Fig. \ref{fig:EnsembleArchitecture}, the approach followed is to sum the output of the Softmax layer of the DL models and determine the output from that sum. As it was previously mentioned in section \ref{sec:Architecture}, the last layer of each DL model in this work is going to perform a Softmax function which determines the likelihood of belonging to a particular class. Due to the fact that all models show their output in the same way and scale, the outputs of each the models in the ensemble was added and the output class was the class which had the highest value in the ensemble.

\section{Tests} \label{sec:test}
In this paper, two sets of tests are presented to analyse the influence of using additional signals to score the sleep stages of a PSG. The first set of tests will explore the influence of use as input more signals and its impact on the classification. Additionally, these first tests were also used to measure the influence of combining more patients during training in the batches. After that set, this paper goes a step further and it also explores the impact of those compose models when they are used in an ensemble schema instead of simple ones. The aim of that last set is to identify if the models with more than one signal can identify additional information which cannot be capture with the ensemble of simple models. 

\begin{table}[h]
    \caption{Training parameters}
    \label{tab:parameters}
    \centering
    \scriptsize
    \begin{tabular}{c|c}
\hline
         \textbf{Parameter}  & \textbf{Value}\\ \hline
         Optimizer  & ADAM \\
         Loss       & cross-entropy \\
         Learning rate & $10^{-4}$\\
         Iterations & 100\\
         Batch size & 32 \\
         Early stopping & 10\\ \hline
    \end{tabular}
\end{table}

For the test, the SHHS-1 dataset was split into 3 subsets for training, validation, and test. Two points should be highlighted about this division, first, the split process was done according to the number of patients instead of the patterns or sections. The reason is to perform what is called a patient-record division in order to keep the 3 different datasets as much separated as possible and closer to a real scenario. Second, the division was performed with a hold-out schema according to 70:10:20 proportion for each one of the three subsets: training, validation, and test, respectively. These subsets were saved and reused in each of the tests of this paper in order to ensure the comparability of the results.

\begin{table}[h]
    \caption{Sleeping Stages on test}\label{tab:testData}
    \centering
    \scriptsize
    \begin{tabular}{ccccc c r}
    \hline
    Awake & N1 & N2 & N3 & REM & &TOTAL\\
    512,112 &65,508& 712,330&227,090&243,183 & &1,760,223\\
    (29\%)&(4\%)&(40\%)&(13\%)&(14\%)& & \\
    \hline
    \end{tabular}
\end{table}

Once the datasets were defined, a series of experiments were carried out following always the same pipeline with the parameters described in Table \ref{tab:parameters}. All tests shown in this paper have been executed in the same machine, an Intel i7 2.6GHz with 32Gbytes of RAM and the support of an NVidia Titan V graphical card. To run each experiment, i.e. run the training and test, the described machine has taken between 2 and 8 days. 

\begingroup
\setlength{\tabcolsep}{6pt} 
\renewcommand{\arraystretch}{1.5}
\begin{table}[h]
    \caption{Naming of the models according to the signals used as input}
    \label{tab:Model_names}\centering
    \scriptsize
    \begin{tabular}{lccc}
         & \multicolumn{3}{c}{Input Signals}\\
        Model Name  & C3A2 & C4A1 & EMG \\
        \cline{2-4} 
        C3A2 & $\star$ &      &      \\
        C4A1 &     &$\star$ &      \\
        EMG  &     &      &$\star$ \\
        EEGs &$\star$&$\star$ &    \\
        C3A2\_EMG & $\star$&     & $\star$ \\
        C4A1\_EMG &       &$\star$&$\star$ \\
        EEG\_EMG & $\star$&$\star$ &$\star$ \\
    \end{tabular}
\end{table}
\endgroup

Starting from the same architecture, a training process was performed in mini-batches of 32 patterns. These training steps used as input patterns 5 consecutive sections of 30 seconds sampled at 125Hz, as it was described in section \ref{sec:Dataset}, and the label of the section in the middle as the desired output of the system. The training is performed by calculating the error of the mini-batches with the cross-entropy as loss function, this measures the difference between the output of the network which uses a Softmax function and the desired output codified as a one-hot encoded. In order to reduce the error, the adjustment is performed through a gradient optimisation technique called ADAM \cite{kingma2014adam} with an initial learning rate of $10^{-4}$. It is also worth to be mention that previous to the last layer, which is the only fully-connected layer, it was set a Dropout layer with a probability of 0.5 in order to improve the generalisation. A threshold of 100 iterations in training was set, after each of which the validation dataset was evaluated. The validation dataset was used to set an early stop threshold if after 10 training cycles the loss of the validation did not improve. The system would return the model with the lowest validation error and this is the one which is going to undertake the evaluation with the test dataset. As previously mentioned, the test dataset was composed by 1161 patients previously seen by the network neither on training nor the validation. In Table \ref{tab:testData}, a summary of the sections contained in the test is presented along with the percentage of each particular label with respect to the total.

\begingroup
\setlength{\tabcolsep}{6pt} 
\renewcommand{\arraystretch}{1.5}
\begin{table}[h]
    \caption{F1 for the each test performed changing the number of patients used to build the mini-batches and the input signals}
    \label{tab:Results_Model_Signals}\centering
    \scriptsize
    \begin{tabular}{lcccccccc}
        Model &\multicolumn{8}{c}{Number of Patients} \\
Name            &  1       &  2        &  3       &  4       &  5       &  6       & 7       & 8      \\
\cline{2-9}\\
C3A2            &  0.7101 &  0.6760  &  0.6794 &  0.7255 &  0.7295 & 0.7229 & \textbf{0.7321} & 0.7208\\
C4A1            &  0.7146 &  0.7233  &  0.7159 &  0.7337 &  \textbf{0.7445} &  0.7426 &  0.7327 & 0.7395\\
EMG             &  0.3836 &  0.3596  &  0.4367 &  0.4270 &  0.4429 &  0.4271 &  0.4441 & \textbf{0.4497}\\
EEGs            &  0.6978 &  0.7095  &  0.7020 &  0.7337 &  0.7415 &  \textbf{0.7455} & 0.7441 & 0.7386\\
C3A2\_EMG       &  0.7290 &  0.6762  &  0.6818 &  0.7282 &  \textbf{0.7407} &  0.7170 & 0.7302 & 0.7312\\
C4A1\_EMG       &  0.7235 &  0.7388  &  0.7115 &  0.7547 &  0.7086 &  0.7343 &  0.7494& \textbf{0.7581}\\
EEG\_EMG        &  0.7008 &  0.6911  &  0.7170 &  0.7283 &  0.7434 &  \textbf{0.7540} & 0.7423 & 0.7391\\
    \end{tabular}
\end{table}
\endgroup

Following the described process, a set of experiments were executed. These experiments used the same architecture described in the section \ref{sec:Architecture}, while they only change the number of inputs between the 3 possibilities as it can be seen in Table \ref{tab:Model_names}. For each of those combination of signals, different training processes using alternative number of individuals to compose the mini-batches were also explored and F1 results are presented in Table \ref{tab:Results_Model_Signals}. In that last Table, the best model for each combination of signals is bold. The reason behind also testing the number of patients in the batches is to identify if there is any influence in the training. Due to the fact of the big size of the data, the whole dataset cannot fit into memory and it has been studied if this has any effect in the training by changing the number of patients between 1 to 8, which is the biggest size that the computer used on the test can handle. 

\begin{figure}[ht]
\begin{subfigure}{0.5\textwidth}
\includegraphics[width=1.0\linewidth, scale=0.3]{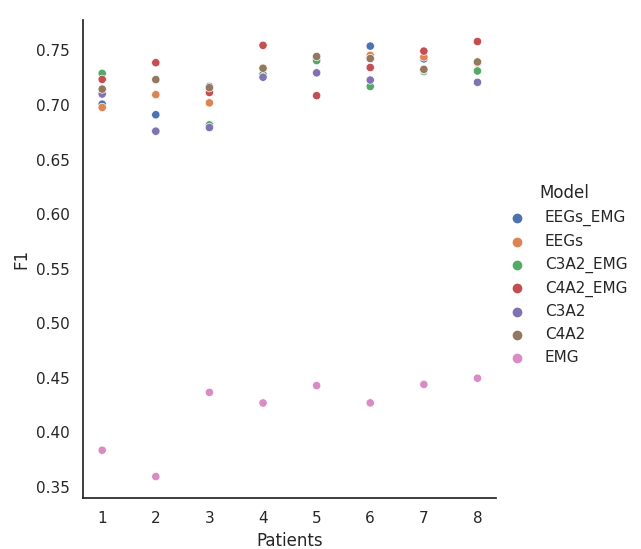} 
\end{subfigure}
\begin{subfigure}{0.5\textwidth}
\includegraphics[width=1.0\linewidth, scale=0.3]{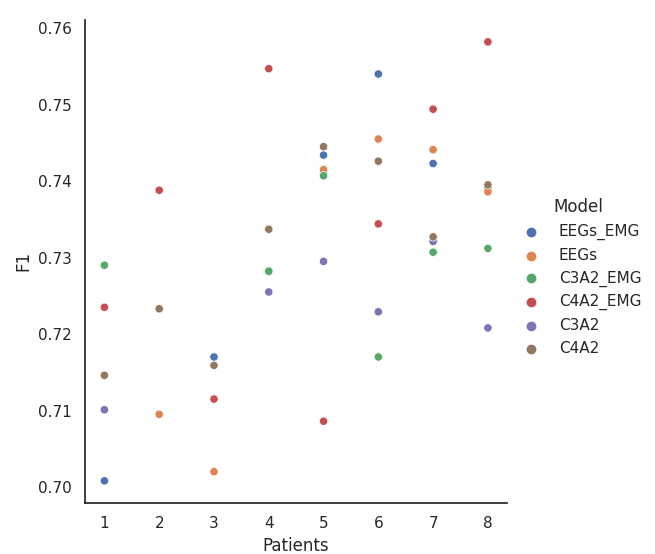}
\end{subfigure}
\caption{Performance of the models attending to $F_{1}$. (Left) Representation of all models. (Right) detail of only the models with at least an $F_{1}$ of 0.70 }
\label{fig:Performance_Models}
\end{figure}

Even though the single-signal models have also good results, the results point out to the improvement of the multi-signal models. When we compare the multi-signal model with the models that use one of the signals used as the input, the result is better in the majority of the cases. Another point that needs also to be highlighted is the fact that, when additional inputs are incorporated into the model, mixing sections from various patients results essential for the generalisation of the model. This last point is especially evident in Figure \ref{fig:Performance_Models} where each model is represented according to its $F_1$ and the number of patients combined in the training mini-batches. That figure exhibits an increasing  Although there is not a clear winner between single-signal and multi-signal models, the data clearly show that without the exception of the model that uses the signal C4A1 the multi-signal ones are usually better especially as more patients are combined in the mini-batches. In that same figure, the plot on the right shows the same information that the one on the left but removing the models that perform less than 0.70 in $F_1$ to improve the readability. It can be seen an increasing tendency on the multi-signal models while the C4A1 model stabilises its progression since 5 patient. 

\begin{figure}[h]
    \centering
    \includegraphics[width=1.0\textwidth]{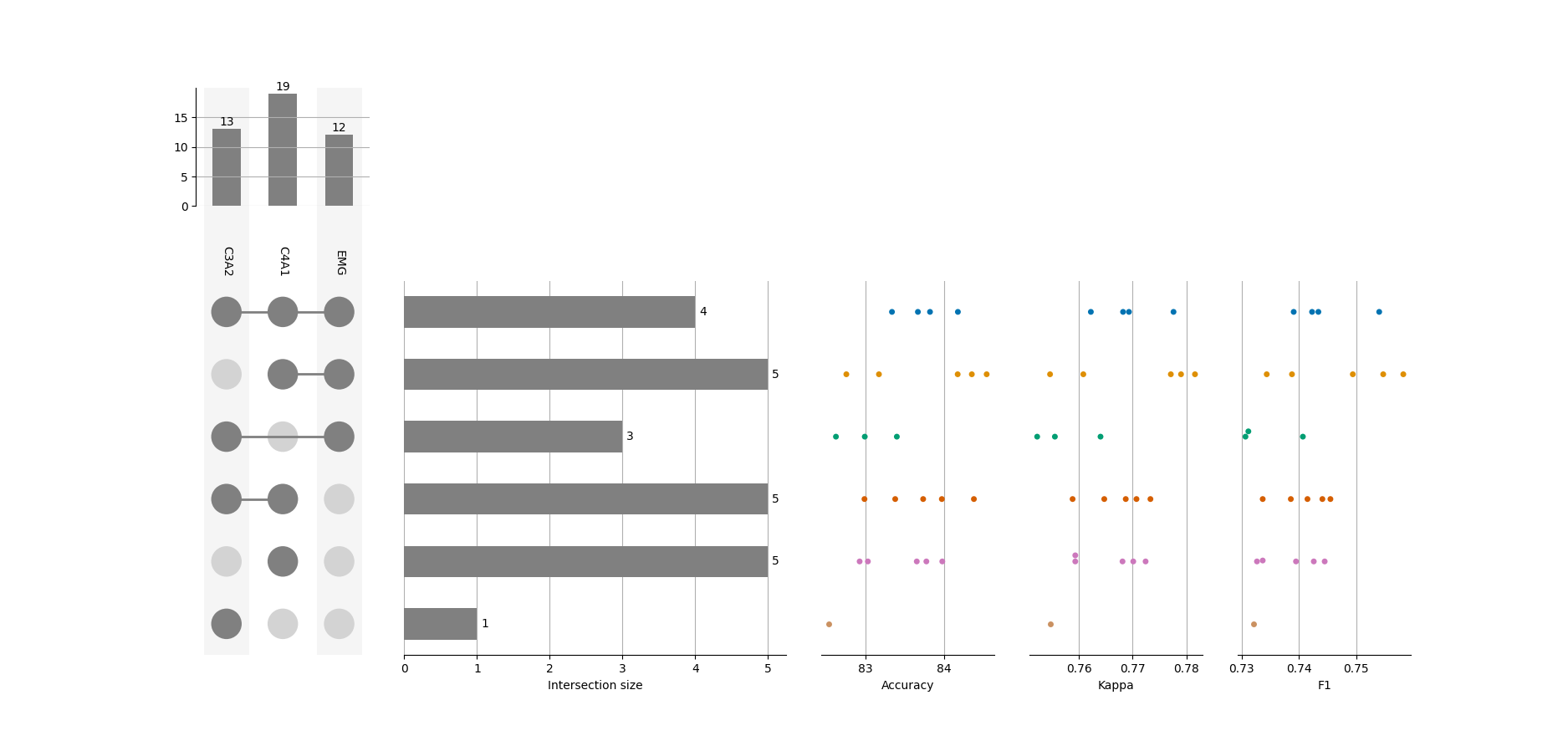}
    \caption{Upset plot of the models achieving at least 0.73 in $F_1$}
    \label{fig:upsetplot}
\end{figure}

\begingroup
\setlength{\tabcolsep}{6pt} 
\renewcommand{\arraystretch}{1.5}
\begin{table}[ht]
    \centering
    \scriptsize
    \caption{Top 10 overall models in test}
    \label{tab:Best_models_overall}
    \begin{tabular}{ccc ccc ccc}
    \hline
    \multicolumn{3}{c}{Input Signals} & &Number of& & \multirow{2}{*}{Accuracy} & \multirow{2}{*}{$\kappa$} & \multirow{2}{*}{$F_1$}\\
        C3A2 & C4A1&EMG& &Patients& & & &(macro)\\
        \cline{1-3} \cline{5-5} \cline{7-9}\\
               &$\star$&$\star$&& 8 && 84.54\%&  0.7816 & 0.7581\\
               &$\star$&$\star$&& 4 && 84.35\%&  0.7790 & 0.7547\\
        $\star$&$\star$&$\star$&& 6 && 84.17\% & 0.7776 & 0.7540\\
               &$\star$&$\star$&& 7 && 84.16\%&  0.7771 & 0.7494\\
        $\star$&$\star$&       && 6 && 84.37\% & 0.7707 & 0.7455\\
               &$\star$&       && 5 && 83.97\% & 0.7723 & 0.7445\\
        $\star$&$\star$&       && 7 && 83.97\% & 0.7733 & 0.7441\\
        $\star$&$\star$&$\star$&& 5 && 83.66\% & 0.7681 & 0.7434\\
               &$\star$&       && 6 && 83.77\% & 0.7701 & 0.7426\\
        $\star$&$\star$&$\star$&& 7 && 83.82\% & 0.7693 & 0.7423\\
    \end{tabular} 
\end{table}
\endgroup

On the Figure \ref{fig:upsetplot}, the models which score at least 0.73 in the $F_1$ are represented. This threshold was set considering the average of the models which perform over 0.7 and they could be considered comparative to the actual state of the art. A closer look to the information on that figure shows two key points. First, most of the models on the table are multi-signal models. In fact, if a we put the spotlight on the top 10 models represented in Table \ref{tab:Best_models_overall}, the multi-signal models achieve 8 out of 10 in that table which ranks the models according to their performance in $F_1$ along with the Accuracy and $\kappa$. The second point to highlight is that C4A1 seems to have more information in this dataset while, in most of the single-signal models of the literature, both EEGs have been indistinctly used. The models using this signal systematically obtain better results and it is in most of the best models as can be seen in Figure \ref{fig:upsetplot} and it is also on all top models in Table \ref{tab:Best_models_overall}.  

\begingroup
\setlength{\tabcolsep}{6pt} 
\renewcommand{\arraystretch}{1.5}
\begin{table}[ht]
    \centering
    \scriptsize
    \caption{Best Models for each input combination}
    \label{tab:Best_Models_Combination}
    \begin{tabular}{cccccccc}
        \hline
        \multicolumn{3}{c}{Input Signals} && \multirow{2}{*}{Accuracy} & \multirow{2}{*}{$\kappa$} & \multirow{2}{*}{$F_1$} & Trainable\\
        C3A2 & C4A1&EMG && & &(macro) & Parameters\\
        \cline{1-3} \cline{5-8}\\
        $\star$ &      &      && 82.54\% & 0.7546 & 0.7321 & 16,782\\
             &$\star$ &      && 83.97\% & 0.7723 & 0.7445& 16,782\\
                  &      &$\star$ &&  58.74\% & 0.3756 & 0.4429& 16,782\\
        $\star$&$\star$ &      && 84.37\% & 0.7707 & 0.7455 & 16,799\\
        $\star$&      &$\star$ && 83.40\% & 0.7640 & 0.7407 & 16,799\\
               &$\star$&$\star$ &&  84.54\% & 0.7816 & 0.7581& 16,799\\
        $\star$&$\star$ &$\star$ && 84.17\% & 0.7776 & 0.7540 & 16,816\\
    \end{tabular}
\end{table}
\endgroup

In Table \ref{tab:Best_Models_Combination},  the best models for each input combination can also be found. It shows how the inclusion of new signals in the input always improve the best result of the single-signal model in Accuracy, $\kappa$ and $F_{1}$ with a minimum additional cost in terms of resources needed by the model. The multi-signal models have the capability to extract features which relate to both signals instead of only one signal which is not possible is to improve the results from a better quality data, i.e., the model with C3A2 and EMG signals do not improve the model which uses C4A1 alone, however, it significantly improves the results from C3A1 model and EMG model. 

\begin{figure}[ht]
\centering
    \begin{subfigure}{0.3\textwidth}
        \includegraphics[width=1\linewidth]{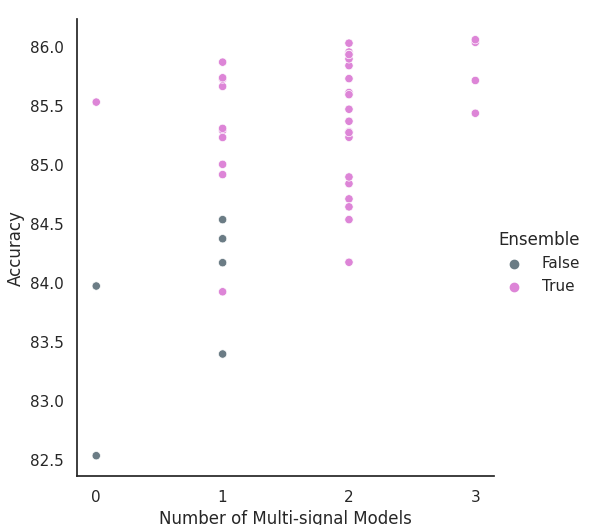} 
    \end{subfigure}
    \begin{subfigure}{0.3\textwidth}
        \includegraphics[width=1\linewidth]{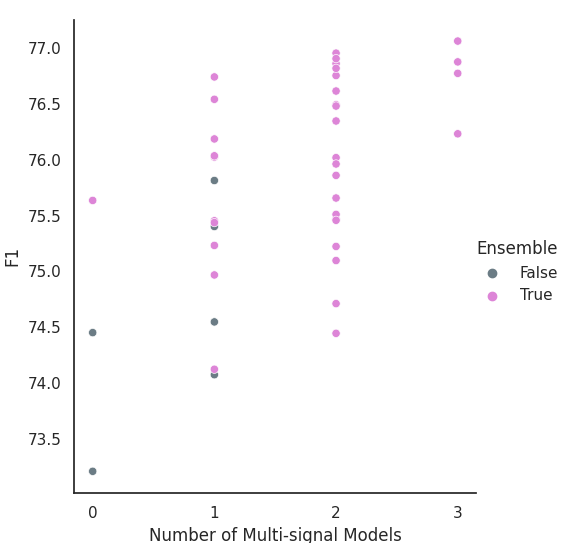}
    \end{subfigure}
    \caption{ Compare the performance of the best models and the different ensembles. The model using only EMG signal has been removed to improve readability due to its poor performance.}
    \label{fig:ensemble_models}
\end{figure}

Moreover, following the ensemble approach described in Section \ref{sec:ensemble}, all possible combinations of ensembles between the best models in Table \ref{tab:Best_Models_Combination} were tested. The aim of these second set of tests is to check if the multi-signal models are able to capture additional information that single-signal models cannot. In figure \ref{fig:ensemble_models}, with the exception of the model only using EMG signal due to its very poor performance, the best single-signal models are represented with the ensembles of all the models presented in Table \ref{tab:Best_Models_Combination}. It is also clear that when the multi-signal models are included the performance in Accuracy and F1 is improved. The top 10 ensemble models can be seen on Table \ref{tab:Best_Ensembled} together with the results for the ensemble of the single-signal models and the best single-signal model as reference.

\begingroup
\setlength{\tabcolsep}{6pt} 
\renewcommand{\arraystretch}{1.5}
\begin{table}[ht]
    \centering
    \scriptsize
    \caption{Top 10 multi-signals ensemble models compared to single-signal ensemble, best multi-signal model and best single-signal model}
    \label{tab:Best_Ensembled}
    \begin{tabular}{ccc c ccc}
        \hline
        \multicolumn{3}{c}{Models} && \multirow{2}{*}{Accuracy} & \multirow{2}{*}{$\kappa$} & \multirow{2}{*}{$F_1$}\\
        Model 1 & Model 2&Model 3 && & &(macro)\\
        \cline{1-3} \cline{5-7}
         EEGs & C4A1\_EMG & EEG\_EMG && 86.06\% & 0.8022 & 0.7707\\
         C3A2 & C4A1\_EMG  & EEG\_EMG  && 85.84\% & 0.8003  &0.7696\\
         C4A1 & C4A1\_EEGs & EEG\_EMG && 85.73\% & 0.7982 & 0.7691\\
         EEGs & C3A2\_EMG & C4A1\_EMG && 86.04\% &  0.8012 &  0.7688\\
         C4A1 & C3A2\_EMG & C4A1\_EMG && 85.90\% &  0.7998 &  0.7686\\
         C4A1 & C3A2\_EMG & EEG\_EMG && 85.93\% &  0.8008 &  0.7682\\
         C3A2\_EMG & C4A1\_EMG & EEG\_EMG && 85.72\% &  0.7979 &  0.7678\\
         C3A2 & EEGs & C4A1\_EMG && 86.03\% &  0.8016 & 0.7676\\
         C3A2 & C4A1 & C4A1\_EMG && 85.87\% &  0.7999 &  0.7674\\
         C4A1 & EEGs & EEG\_EMG && 85.94\% &  0.8001 &  0.7661\\
          & ... & &  & & ...&  \\
         C3A2 & C4A1 & EMG && 85.53\% & 0.7930 & 0.7564\\
         C4A1\_EMG & &  && 84.54\% & 0.7816 & 0.7581\\
         C4A1 & & && 83.97\% & 0.7724 & 0.7445\\
    \end{tabular}
\end{table}
\endgroup

Therefore, the best models include at least one model with multi-signal and, in particular, with a heterogeneous mixture of signals. In fact, when the results are compared to the ensemble of single-signal models, it is clear that at least the mixture of several signals improve the classification as it is shown on the two best models which have all three members of the ensemble as multi-signal models. Additionally, analysing the confusion matrix, Table \ref{tab:confusion_matrix_ensemble} of the best ensemble it is clear that is able to be a very robust and precise model compared to the single models in most of the classes except with the class N1 which is quite underrepresented in the dataset and even the physicians have problems to identify it, too.

\begingroup
\setlength{\tabcolsep}{6pt} 
\renewcommand{\arraystretch}{1.5} 
\begin{table}[ht]
    \caption{Confusion matrix of the best ensemble model in test}\label{tab:confusion_matrix_ensemble}
    \centering
    \scriptsize
       \begin{tabular}{c r ccccc c}
    & &\multicolumn{5}{c}{Ground truth}\\
    & &Awake & N1 & N2 & N3 & REM & \multicolumn{1}{|c}{Precision}\\
    \multirow{5}{*}{\rotatebox[origin=c]{90}{Network}} & Awake & 479,551 & 19,563  &  24,664 &  769  &   9,861 & \multicolumn{1}{|c}{89.86\%}\\
        & N1 & 3,868 &  14,872  &  5,197  &  3    &   1,958 &  \multicolumn{1}{|c}{57.42\%}\\
        & N2 & 20,392 & 20,159  &  635,514 & 50,331  & 22,197 &  \multicolumn{1}{|c}{84.89\%}\\
        & N3 & 1,273  & 12  & 22,544 &   175,868  & 99 &  \multicolumn{1}{|c}{88.02\%}\\
        & REM & 7,028 &  10,902 & 24,411  & 119   & 209,068 &  \multicolumn{1}{|c}{83.12\%}\\
        \cline{2-7}
        &Recall& 93.64\% & 22.70\% & 89.22\% & 77.44\% & 85.97\%\\
    \end{tabular}
\end{table}
\endgroup

Overall, the general performance of the network can be seen in Table \ref{tab:comparison} together with the other works which has used the same dataset. It should be highlighted that the presented work, oppositely to the one presented by \cite{sors2018convolutional}, do not perform any additional filtering or clipping of the signals and different to  \cite{Pathak2021STQS:Scoring} do not treat the imbalanced by weighting the training. 

\begingroup
\setlength{\tabcolsep}{8pt} 
\renewcommand{\arraystretch}{1.5} 
\begin{table}[ht]
    \caption{Performance of the different works on SHHS-1 and their complexity}\label{tab:comparison}
    \centering
    \scriptsize
       \begin{tabular}{rc rp{1cm}p{1cm}c cccr@{,}l}
    &Year & Number of & Input & Method&Evaluation&\multirow{2}{*}{Accuracy} & \multirow{2}{*}{$\kappa$} & \multirow{2}{*}{$F_1$}& \multicolumn{2}{c}{Trainable} \\
    & & Patients & & & & & & (macro) & \multicolumn{2}{c}{Parameters}\\\cline{2-11}
    \cite{Biswal2018Expert-levelNetworks}&2018&5804 & 2EEG&	CNN-RNN-RC & 90–10& 77.91\% &0.73 &\\
    \cite{sors2018convolutional}& 2018	& 5,728 & 1EEG& CNN	& 50-20-30	& 87.00\% & 0.81 & 0.86 & 199,068&478\\
    \cite{Fernandez-Blanco2020EEGStage}&2020& 5,804 &2EEG & DS-CNN& 70-10-20 & 85.22\% & 0.79 & 0.76 & 16&799\\
    \cite{Pathak2021STQS:Scoring} & 2021& 5,793 & 2EEG 2EOG 1EMG& CNN-BiLSTM-RC& 81-9-10 & 85.00\%& 0.79&0.76 & $\approx$98&000\\
    \textbf{This work} && 5,804 & 2EEG 1EMG & Ensemble of DS-CNN & 70-10-20& 86.06\%&0.80 & 0.77 & 50&914\\
    \end{tabular}
\end{table}
\endgroup

Even though the model has significantly more parameters, its performance is equal o better to the state of the art. Only \cite{sors2018convolutional} presents better results, however, this work removes some patients from the dataset and clips the signals in order to remove the periods of awake previous to the start of the sleep and it has treated the imbalance during training. Therefore the results could not be exactly reproduced, although it should be mentioned that the model presented in this paper C4A1 uses a similar architecture while using the same datasets for train, validate and test than this one. The results of that model are on Table \ref{tab:Best_Ensembled} and, as it was mentioned the proposed ensemble model improves those values. Moreover, comparing the precision in Table \ref{tab:Compare_precision}, it shows this work is the best to deal with the class N1 while showing high values on the remaining classes.it should be mentioned that the precision per class in \cite{Biswal2018Expert-levelNetworks} was not reported, so it was taken out of the comparison. 

\begingroup
\setlength{\tabcolsep}{8pt} 
\renewcommand{\arraystretch}{1.5} 
\begin{table}[ht]
    \caption{Comparing precision per class}\label{tab:Compare_precision}
    \centering
    \scriptsize
       \begin{tabular}{r ccccc}
    &Awake&N1&N2&N3&REM\\
    \cline{2-6}
    \cite{sors2018convolutional}& 91.0 & 42.7 & 87.9 & 85.0 & 85.4\\
    \cite{Fernandez-Blanco2020EEGStage}& 91.7 & 54.0 & 84.5 & 83.6 & 87.6\\
    \cite{Pathak2021STQS:Scoring} & 92.5 & 40.3 & 84.4 & 76.0 & 89.1\\
    \textbf{This work} & 89.7 & 57.4 & 84.9 & 88.0 & 83.1\\
    \end{tabular}
\end{table}
\endgroup

\section{Conclusions}\label{sec:conclusions}
Several conclusions can be extracted from this work, first and foremost, a new approach based on an ensemble of PS-CNN improve how to tackle the problem of scoring the sleep stages. The proposed model has shown to be able to identify the less frequent class better than any previous approach while it keeps similar percentages in the remaining classes. Different from previous approaches such as \cite{sors2018convolutional,Pathak2021STQS:Scoring}, this point has been achieved without any additional imbalance treatment, as seen on table \ref{tab:Compare_precision}. Moreover, the proposed approach shows, on the authors' best known, the best results on this dataset up to date, without trimming the signal or removing patients.

Additionally, it is also worth mentioning the deep study of the different combinations of signals which has to draw two main conclusions. First, the inequality of the EEGs recorded in the dataset. As it can be seen in Table \ref{tab:Results_Model_Signals}, the channel C4A1 seems to have more information alone than the other two signals while using the same datasets. However, the second conclusion is that this information can be complemented with additional signals, which allow to mine information that relates the two signals in a single convolutional model. Even with the EMG which has shown the poorest performance of all signals alone, when it is combined with one of the EEGs channels the results are significantly better than the results of models using each EEG alone.

Those models were later used to compose different ensembles, and those ensembles have shown how the use of multiple signals, especially when they are heterogeneous, has better results than the ensemble of single-signals models of any models alone.

Finally, it should also be mentioned that this type of ensemble is possible due to the use of models which use Separable Convolutional layers. The reduction in the number of parameters allows the composition, train and use of this kind of model not only in high-end infrastructures but in modest computers. For example, as it can be seen on table \ref{tab:comparison} a similar ensemble with the model proposed in \cite{sors2018convolutional} would require no less than 600 million parameters while the proposed here are roughly 51,000.

\section{Futures works}\label{sec:future}
From this work, there are several lines which can be explored. Going along with the same dataset, the SHHS-1 has several signals more, such as electrooculograms or electrocardiograms. However, these other signals have a different sample ratio than the EEG ad EMG, therefore, exploring the best way to incorporate that information in the classification models could be the next step. Although some recent works such as \cite{andreotti2018multichannel} have shown their advantages on small datasets. Several are the possibilities in this sense, from the inclusion of specific models for signals with a different ratio in an ensemble model to the incorporation of that signals in a Deep Learning model.

Second, another point that should be explored, which is related to the previous one, is how to treat the missing values. Some recent works have started to dig into this question, such as \cite{Pathak2021STQS:Scoring}, performing an evaluation in the time domain based on Bidirectional Long Short Term Memory (Bi-LSTM) architecture. However, other possibilities should be explored from the use of Generative Adversarial Networks to a simply interpolation.

Third, another point to be improved is the problem that this dataset has with the class N1 which is significantly underrepresented in the whole dataset. Although the presented model represents an improvement over the state-of-the-art, dealing with this problem could be another object of study. For example, Weighting during training the mistakes classified of the less represented could be one of the approaches to compare with other.

Fourth, although the presented model shows a significant performance on the SHHS-1, other works have used different channels or signals from other datasets. How the transfer learning \cite{shin2016deep, rebuffi2018efficient} would work in this kind of signals, using a different and smaller dataset such as\cite{o2014montreal}, is a question to check.

Finally, the issue of the inconsistent and unreliable ground truth should also be tackle. A schema such as the ones recently presented like the man-in-the-loop \cite{Zanzotto2019Viewpoint:Intelligence} architecture could be a possible approach. In fact, this should not be focused exclusively on the improvement of the performance but the identification of the right set features to label the signals instead of the expertise.

\section*{Acknowledgements}
The authors want to acknowledge the support from NVdivia, who has  donated the GPU used in the experiments of this publication, and Centro de Supercomputación de Galicia (CESGA), who allows to conduct the first exploratory tests on their installations.
\section*{Funding}
This work is partially supported by Instituto de Salud Carlos III, grant number PI17/01826 (Collaborative Project in Genomic Data Integration (CICLOGEN) funded by the Instituto de Salud Carlos III from the Spanish National Plan for Scientific and Technical Research and Innovation 2013–2016 and the European Regional Development Funds (FEDER)—“A way to build Europe.”. It was also partially supported by different grants and projects from the Xunta de Galicia [ED431D 2017/23; ED431D 2017/16; ED431G/01; ED431C 2018/49; IN845D-2020/03]. Finally, another source of support was the CYTED network (PCI2018\_093284) funded by the Spanish Ministry of Innovation and Science.

\bibliographystyle{unsrt}
\bibliography{bibliography.bib, mendeley-references.bib}

\end{document}